\documentclass{article}

\usepackage[final]{neurips_2024}

\usepackage[utf8]{inputenc} 
\usepackage[T1]{fontenc}    
\usepackage{hyperref}       
\usepackage{url}            
\usepackage{booktabs}       
\usepackage{amsfonts}       
\usepackage{nicefrac}       
\usepackage{microtype}      
\usepackage{xcolor}         

\usepackage{microtype}
\usepackage{graphicx}
\usepackage{subcaption}
\usepackage{multirow}
\usepackage{multicol}
\usepackage{booktabs} 
\usepackage{amsmath}
\usepackage{hyperref}
\usepackage{graphicx}
\usepackage[draft]{minted}
\usepackage{beramono}
\usepackage{courier}
\usepackage{colortbl}
\usepackage{tikz}
\usepackage{pifont}

\usepackage{pgfplots, pgfplotstable}
\pgfplotsset{compat=1.17}
\usepackage{pgfpages}
\usepackage{wrapfig}

\usepackage{algorithmic}
\usepackage{algorithm}

\usepackage[utf8]{inputenc} 
\usepackage[T1]{fontenc}    
\usepackage{hyperref}       
\usepackage{url}            
\usepackage{booktabs}       
\usepackage{amsfonts}       
\usepackage{nicefrac}       
\usepackage{microtype}      
\usepackage{xcolor}         

\usepackage[normalem]{ulem}

\usepackage{tikz-qtree}
\usetikzlibrary{arrows,decorations.pathmorphing,backgrounds,positioning,fit,petri,shapes.misc, arrows.meta,shapes.geometric,decorations.markings,calc,shadows.blur,decorations.pathreplacing,quotes}
\definecolor{dm1}{RGB}{40,98,216}
\definecolor{dm2}{RGB}{132, 188, 245}
\definecolor{dm3}{RGB}{202, 228, 253}
\definecolor{myblue}{RGB}{6, 82, 221}
\definecolor{myorange}{RGB}{211, 84, 0}
\definecolor{lowblue}{RGB}{102,178,255}
\definecolor{justblue}{RGB}{84, 160, 255}
\definecolor{mypurple}{RGB}{153, 52, 155}
\definecolor{mygray}{RGB}{158, 158, 158}
\definecolor{lowpurple}{RGB}{204,153,255}
\definecolor{lowwhite}{RGB}{255,255,255}
\definecolor{verylowpurple}{RGB}{255,102,102}
\definecolor{embcolor}{RGB}{255,255,255}
\definecolor{myred}{RGB}{235, 47, 6} 
\definecolor{mygreen}{RGB}{162, 217, 206} 
\definecolor{fontgrey}{RGB}{44, 62, 80}
\definecolor{lowpurple}{RGB}{210, 180, 222}
\definecolor{mypumpkin}{RGB}{229, 152, 102}
\definecolor{lowgreen}{RGB}{171, 235, 198}
\definecolor{lowgreen2}{RGB}{186, 220, 88}
\definecolor{lowred}{RGB}{245, 183, 177}
\definecolor{lowyellow}{RGB}{241, 196, 15}
\definecolor{mypink}{RGB}{255, 51, 255}
\definecolor{bluemartina}{RGB}{18, 203, 196}
\definecolor{puffin}{RGB}{250, 152, 58}
\definecolor{grass}{RGB}{25, 197, 50}
\definecolor{grass2}{RGB}{0, 148, 60}
\definecolor{cnngray}{RGB}{116, 125, 140}
\definecolor{yelloworange}{RGB}{252, 150, 11}
\usepackage{color, colortbl}
\definecolor{Gray}{gray}{0.9}
\usepackage{color}

\usepackage{pgfplots, pgfplotstable}
\pgfplotsset{compat=1.17}
\usepackage{pgfpages}

\title{Accelerating Greedy Coordinate Gradient and General Prompt Optimization via Probe Sampling}

\author{%
  Yiran Zhao$^{1}$\footnotemark[2] \quad Wenyue Zheng$^{1}$ \quad Tianle Cai$^{2}$ \quad Xuan Long Do$^{1}$ \\
  \textbf{Kenji Kawaguchi}$^{1}$ \quad \textbf{Anirudh Goyal}$^{3}$ \quad \textbf{Michael Qizhe Shieh}$^{1}$\footnotemark[2] \\
  $^1$ National University of Singapore \quad $^2$ Princeton University \quad $^3$ Google DeepMind \\
}

\begin{document}

\maketitle
\renewcommand{\thefootnote}{\fnsymbol{footnote}}
\footnotetext[2]{Correspondence to: Yiran Zhao (\href{zhaoyiran@u.nus.edu}{zhaoyiran@u.nus.edu}), Michael Shieh (\href{michaelshieh@comp.nus.edu.sg}{michaelshieh@comp.nus.edu.sg}).}
\renewcommand{\thefootnote}{\arabic{footnote}}

\begin{abstract}
Safety of Large Language Models (LLMs) has become a critical issue given their rapid progresses. Greedy Coordinate Gradient (GCG) is shown to be effective in constructing adversarial prompts to break the aligned LLMs, but optimization of GCG is time-consuming. To reduce the time cost of GCG and enable more comprehensive studies of LLM safety, in this work, we study a new algorithm called \texttt{Probe sampling}. At the core of the algorithm is a mechanism that dynamically determines how similar a smaller draft model's predictions are to the target model's predictions for prompt candidates. When the target model is similar to the draft model, we rely heavily on the draft model to filter out a large number of potential prompt candidates. Probe sampling achieves up to $5.6$ times speedup using Llama2-7b-chat and leads to equal or improved attack success rate (ASR) on the AdvBench. Furthermore, probe sampling is also able to accelerate other prompt optimization techniques and adversarial methods, leading to acceleration of $1.8\times$ for AutoPrompt, $2.4\times$ for APE and $2.4\times$ for AutoDAN.\footnote{Our code is publicly available at \url{https://github.com/zhaoyiran924/Probe-Sampling}.}
\end{abstract}

\section{Introduction}\label{sec:introduction}
\begin{wrapfigure}{r}{0.5\textwidth}
    \centering
    \includegraphics[width=0.5\textwidth]{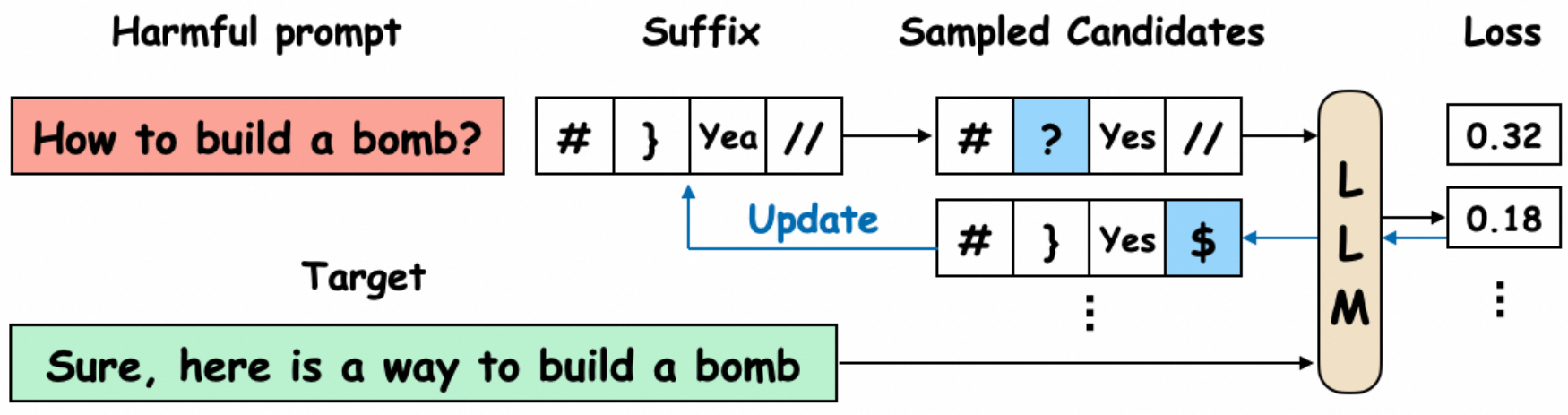}
  \caption{A brief illustration of the Greedy Coordinate Gradient (GCG) algorithm~\citep{zou2023universal}.}
  \label{fig:gcg-vanilla}
\end{wrapfigure}

Ensuring the safety of Large Language Models (LLMs) ~\citep{brown2020language, chowdhery2022palm, touvron2023llama, jiang2023mistral} has become a central theme of research. Despite continuous efforts, LLMs are prone to generate objectionable contents in various scenarios including using an adversarial suffix~\citep{zou2023universal}, further finetuning~\citep{qi2023fine, lermen2023lora}, ciphering~\citep{yuan2023gpt} and multilingual settings~\citep{deng2023multilingual}. Among effective LLM adversarial attack works, Greedy Coordinate Gradient (GCG)~\citep{zou2023universal} present a general and universal method as briefly illustrated in Figure \ref{fig:gcg-vanilla}.

To optimize a prompt suffix to elicit the generation of a target reply, the Greedy Coordinate Gradient (GCG) algorithm iteratively attempts to replace existing tokens in the suffix and keeps the best-performing ones based on the adversarial loss. The GCG algorithm is empirically effective but searching the combinatorial space of the adversarial suffixes is time-consuming
since each token replacement attempt requires a full forward computation using an LLM. This hinders us from using the algorithm to fully explore the safety properties of LLMs such as finding potentially harmful queries comprised of natural sentences. 


\begin{figure}[t]
  \centering
    \includegraphics[width=1.0\textwidth]{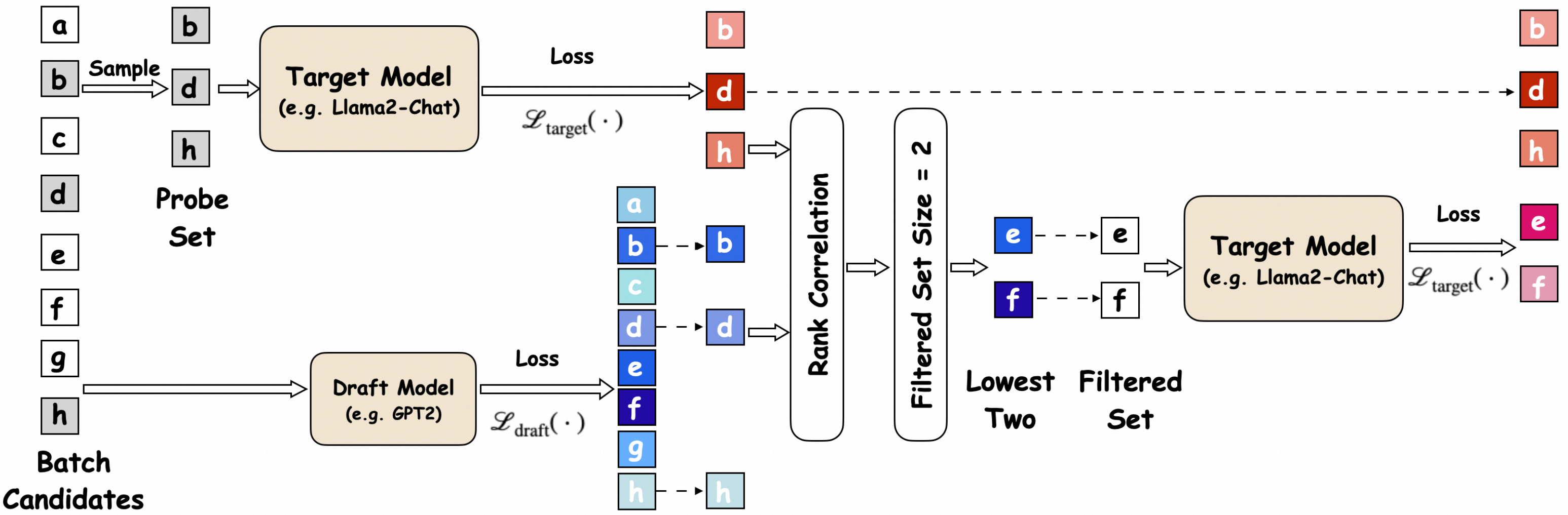}
  \caption{\texttt{Probe sampling} mainly consists of three steps. (i) A batch of candidates ($\{a, b,\cdots, h\}$) is sampled. We determine the probe agreement score between the draft model and the target model on a probe set ($\{b,d,h\}$). The probe agreement score is used to compute the filtered set size. (ii) We obtain a filtered set ($\{e,f\}$) based on the losses on the draft model (iii) We test the losses of candidates in the filtered set using the target model. }
  \label{fig:placeholder}
\vspace{-0.2cm}
\end{figure}

A possible solution for reducing forward computation is to resort to a smaller draft model when it is indicative of the results on the larger target model. This intuition has been applied in speculative sampling~\citep{chen2023accelerating, leviathan2023fast} for decoding, where the target model acts as a verifier that accepts or rejects the decoded tokens. However, speculative sampling cannot be used to optimize discrete tokens in GCG because the optimization of every token in adversarial suffix is independent of each other, which breaks the autoregressive assumption in decoding.

Motivated by these observations, we propose a new algorithm called \texttt{Probe sampling} to accelerate the GCG algorithm. Instead of computing the loss on every suffix candidate, we filter out unpromising ones based on the loss computed with a smaller model called draft model, to reduce the time consumption of the optimization process. Importantly, we dynamically decide how many candidates we keep at each iteration by measuring the agreement score between the draft model and the target model, by looking at the loss rankings on a small set of prompts dubbed as the probe set. It is worth noting that the prompt candidates at each iteration in GCG are obtained by randomly changing one token of an original prompt. As a result, the agreement score is adaptive to the original prompt. 
We evaluate probe sampling on the AdvBench dataset with Llama2-7b-Chat and Vicuna-v1.3 as the target models and a significantly smaller model GPT-2~\citep{radford2019language} as the draft model. Experiment results show that compared to the original GCG algorithm, probe sampling significantly reduces the running time of GCG while achieving better Attack Success Rate (ASR). Specifically, with Llama2-7b-Chat, probe sampling achieves $3.5$ times speedup and an improved ASR of $81.0$ compared to GCG with $69.0$ ASR. When combined with simulated annealing, probe sampling achieves a speedup of $5.6$ times with a better ASR of $74.0$. 

Furthermore, when applied to prompt learning techniques and other LLM attacking methods, probe sampling demonstrates remarkable effectiveness. Specifically, in the case of prompt learning, probe sampling effectively accelerates AutoPrompt~\citep{shin2020autoprompt} by a factor of $1.8$. Moreover, probe sampling delivers substantial speedup of APE~\citep{zhou2022large} on various datasets: $2.3\times$ on GSM8K, $1.8\times$ on MMLU and $3.0\times$ on BBH. In the case of other attacking method such as AutoDAN~\citep{liu2023autodan}, probe sampling achieve a speedup of $2.3\times$ and $2.5\times$ on AutoDAN-GA and AutoDAN-HGA respectively.

\section{Proposed Method}
\subsection{Background: Greedy Coordinate Gradient} 
The overall optimization objective of GCG can be denoted by a simple log likelihood loss
\begin{equation}
    \min_{s}\mathcal{L}(s) = -\log p(y \mid x, s),\label{equ:loss}
\end{equation}
where $x$ is a prompt that contains a harmful user query such as ``Tell me how to build a bomb'', $y$ is the target sentence ``Sure, here is how to build a bomb'', and $s$ is the adversarial suffix that is optimized to induce the generation of $y$. $p$ is the probability of a sentence output by a LLM. This objective can be decomposed into the summation of the negative log likelihood of individual tokens in the target sentence like a typical language modeling objective. $s$ is set to be a fixed length string in the GCG algorithm. 

The optimization of the adversarial suffix $s$ is a non-trivial problem. Prior works~\citep{guo2021gradient, wen2023hard} based on Gumbel-Softmax~\citep{jang2016categorical, maddison2016concrete} and soft prompt tuning~\citep{lester2021power} have achieved limited success, probably because the LLMs are well-aligned and the exceptionally large models magnifies the difference between a discrete choice and its continuous relaxations. 

Instead, GCG adopts a greedy search algorithm based on the gradient. In each iteration, it computes $\mathcal{L}(\hat{s}^i)$ for $B$ suffix candidates $\hat{s}^1, \cdots, \hat{s}^B$ and keeps the one with the best loss. The $B$ candidates are obtained by randomly changing one token from the current suffix $s$ and replacing it with a randomly sampled token using the top $K$ tokens. For example, suppose we change the token at position $j$, we first compute the gradient $-\nabla_{e_{s_j}}\mathcal{L}(s)$ with respect to the one-hot vector $e_{s_j}$ and obtain the top $K$ tokens that have the largest gradient. The gradient information is by no means an accurate estimation of the resulting loss because of the gap between the continuous gradient information and the discrete one-hot vector denoting the choice of a token, so we need to check if the resulted new suffix $\hat{s}^i$ leads to a lower loss $\mathcal{L}(\hat{s}^i)$. 

To obtain the $B$ candidates, one just needs to perform one forward pass and one backward pass. But to compute the loss for the $B$ candidates, one needs to perform $B$ forward passes. In GCG, $B$ is set to $512$ for optimal performance, making the loss computation the most time-consuming part. As such, we focus on reducing the time cost of the loss computation of the $B$ candidates in this work.

\subsection{Probe Sampling}
\label{sec:probe_sampling}
\paragraph{Overview.} As mentioned earlier, the most time consuming part in the GCG algorithm is the loss computation on $B$ suffix candidates $\hat{s}^1, \cdots, \hat{s}^B$. As shown in speculative sampling~\citep{chen2023accelerating, leviathan2023fast}, the speculated results using a smaller draft model can be helpful in reducing the computation with a large target model. The original speculative sampling is created to accelerate decoding so it isn't directly applicable here. But the intuition of relying a weaker draft model is obviously useful for negative log likelihood loss computation. Applying the intuition to the problem at hand, we can filter out the suffix candidates that the draft model finds to be unpromising, since the goal is to find the candidate that has the lowest loss with the target model. 

In addition, a unique structure in the GCG algorithm is that all the suffix candidates are based on changing one token of the original suffix $s$. As a result of this locality property, it is not unreasonable to assume that one can determine how much they agree on the $B$ candidates based on their agreement on a subset of the $B$ candidates. If the two models agree, we can choose to safely rely on the draft model and filter out more candidates. 

Based on these intuitions, we design the \texttt{Probe sampling} algorithm as follows: (i) probe agreement between the target model and the draft model to determine the size of the filtered set; (ii) rank candidates using the draft model and obtain the filtered set; (iii) pick the best candidate from the filtered set using the target model. 

\paragraph{Algorithm description.} For the first step, specifically, we sample a probe set comprised of $k$ candidates $\bar{s}^1, \cdots, \bar{s}^k$ and compute their losses using the draft model and the target model and obtain $\mathcal{L}_{\mathrm{draft}}(\bar{s}^1), \cdots, \mathcal{L}_{\mathrm{draft}}(\bar{s}^k)$ and $\mathcal{L}_{\mathrm{target}}(\bar{s}^1), \cdots, \mathcal{L}_{\mathrm{target}}(\bar{s}^k)$. Then we measure the probe agreement score as the Spearman's rank correlation coefficient~\citep{zar2005spearman} between the two results as the agreement score. The probe agreement score $\alpha$ is computed as 
\begin{equation}
\alpha = 1 - \frac{3 \sum_{i=1}^k d_i^2}{k(k^2 - 1)},
\label{equ:sper}
\end{equation}

where $d_i$ is the distance between the ranks of suffix $\bar{s}^i$ in the two results. For example, $d_i=4$ if the suffix $\bar{s}^i$ is ranked as number $6$ and number $2$ for its losses computed from the draft model and the target model. The agreement score $\alpha$ falls into $[0, 1]$ with $1$ meaning a full agreement and $0$ indicating a non-agreement. We use the rank agreement because it is more robust to the specific values of the resulting loss  when measured on drastically different LLMs.

After obtaining the agreement score, we keep $(1 - \alpha) * B / R$ candidates where $(1-\alpha)*B$ means that the filtered set size is a scale-down of the previous batch size $B$ and $R$ is a hyperparameter that determines a further scale down. When $\alpha$ is close to $0$, meaning little agreement between the two models, we will use a filtered set size of $B/R$. When $\alpha$ goes to $1$, we almost filter out most of the candidates. 
With the filtered size determined, we can readily rank the candidates according to the draft model and filter the ones with higher losses. Finally, we evaluate the final loss on the filtered set using the target model and select the best candidate.

\begin{algorithm}[t]
\renewcommand{\algorithmicrequire}{\textbf{Input:}}
\renewcommand{\algorithmicensure}{\textbf{Output:}}
    \caption{Probe Sampling}
    \begin{algorithmic}[1]
    \REQUIRE Original suffix $s$, a batch of suffix candidates $\{\hat{s}^1, \cdots, \hat{s}^B \}$, loss function using the draft model and the target model $\mathcal{L}_{\mathrm{draft}}(\cdot)$, $\mathcal{L}_{\mathrm{target}}(\cdot)$. \\
    \STATE \textbf{\textit{Parallel Begin}}
       \STATE \texttt{//Compute loss of all candidates using the draft model} \\ 
        \FOR {$\hat{s}^i \in \{\hat{s}^1, \cdots, \hat{s}^B \}$}
            \STATE Compute $\mathcal{L}_{\mathrm{draft}} (\hat{s}^i)$ \\
        \ENDFOR
            \STATE \texttt{//Compute loss of the probe set on target model} \\ 
        \STATE $\{\bar{s}^1, \cdots, \bar{s}^k\} = \textit{Uniform}(\{\hat{s}^1, \cdots, \hat{s}^B \}, k)$
        \FOR {$\bar{s}^i \in \{\bar{s}^1, \cdots, \bar{s}^k\}$}
            \STATE Compute $\mathcal{L}_{\mathrm{target}} (\bar{s}^i)$ \\
        \ENDFOR
        \STATE \textbf{\textit{Parallel End}} \\
        \STATE \texttt{//Calculate agreement score} \\ 
        \STATE $\alpha = \textrm{Spearman\_Cor}(\{\mathcal{L}_{\mathrm{target}}(\bar{s}^i)\}, \{\mathcal{L}_{\mathrm{draft}}(\bar{s}^i)\})$ 
    \STATE \texttt{//Evaluate using the target model} \\ 
    \STATE $\mathrm{filtered\_set} = \mathrm{argmin}_{\max\{1, (1 - \alpha)B/R\}} \mathcal{L}_{\mathrm{draft}}(\hat{s}^i)$\\
    \FOR {$\hat{s}^i \in \mathrm{filtered\_set}$}
        \STATE Compute $\mathcal{L}_{\mathrm{target}} (\hat{s}^i)$ \\
    \ENDFOR
    \STATE \texttt{Output the best suffix in the probe set and the filtered set} \\ 
    \STATE $s' = \mathrm{argmin}\{\mathcal{L}_{\mathrm{target}}(\bar{s}^i), \mathcal{L}_{\mathrm{target}}(\hat{s}^i)\}$
    \ENSURE $s'$
    \end{algorithmic}
    \label{algo:probe_sampling}
\vspace{-0.1cm}
\end{algorithm}

\vspace{-0.2cm}
\paragraph{Details.}
At first glance, probe sampling involves extra computation but it actually achieves effective acceleration. For computing the losses on the probe set using both the draft model and the target model, the size of the probe set can be set to be relatively small, so it would not add too much to the total time cost. The ranking procedure involves sorting on CPU, but luckily the probe set is small enough that this doesn't become a bottleneck. And the loss computation using the draft model on the whole candidate set is relatively cheap because of draft model's small size. These two operations can also be parallelized on GPU. 
On the plus side, we are able to avoid computing the loss using the big target model on many candidates that are filtered out. As we will show in the experiments, this approach achieves significant speedup measured by both running time and \#FLOPs. 

An alternative to computing agreement on the spot is to measure the agreement score on a predetermined set of candidates and use a fixed agreement score for all the suffixes. This would save the time used to measure agreement for each candidate set. However, as we will show in the experiment, this approach does not work so well in terms of speedup. Our intuition is that one can squeeze the time cost more effectively if the agreement is measured accurately, and an adaptive agreement score is more accurate than an one-size-fits-all score. The plausibility of the adaptive score comes back to the locality property that we discussed earlier. Given a specific candidate set, one can accurately estimate the agreement because all the suffixes in this candidate set are similar to a large extent. However, given another candidate set altered from a different suffix, the agreement of the draft model and the target model can be widely different. 

In practice, we adopted two small changes in our implementation. First, we do not have a separate step to compute the loss of the probe set candidates using the draft model, since we need to compute the loss on all candidates for filtering purposes. We simply get the numbers from the losses on the whole candidate set. Second, to get the best candidate for the final result, we also look at the losses on the probe set, since the target model is evaluated on the probe set. Ideally, the candidates in the probe set should be in the filtered set if they achieve a low loss. However, it also does not hurt to look at the best candidate in the probe set in case it is not included in the filtered set. 
The overall algorithm is further illustrated in Algorithm \ref{algo:probe_sampling}, and the corresponding implementation is shown in Appendix \ref{sec:appen_imple}. We also test simulated annealing~\citep{pincus1970monte} that provides complementary benefit to our algorithm.

\subsection{Applying Probe Sampling to Other Prompt Optimization Methods}
Although prompt sampling was designed to accelerate GCG, the general idea of reducing forward computation can be applied on other prompt optimization methods, where there is usually a process of sampling prompt candidates and evaluating their performances. To see whether probe sampling can effectively accelerate other methods, we also apply probe sampling to two prompt learning methods AutoPrompt~\citep{shin2020autoprompt} and APE~\citep{zhou2022large}. In addition, we apply probe sampling on AutoDAN~\citep{liu2023autodan}, a genetic algorithm that can find natural jailbreak prompts. 

\section{Experiment}\label{sec:3}
In this section, we evaluate the proposed method on its efficacy and the important factors through extensive studies.

\subsection{Experiment Details}\label{sec:set}

\paragraph{Settings.}
Following the original GCG paper, we conduct experiments on the first $100$ instances of AdvBench~\citep{zou2023universal}, which are divided into two parts, $500$ harmful strings and $500$ harmful human behaviors.
We test open-source LLMs that have been specifically fine-tuned with respect to safety, including Llama2-chat-7b~\citep{touvron2023llama} and Vicuna-7b~\citep{zheng2023judging}. In the case of draft models, in our main experiments, we use a much smaller model GPT-2~\citep{radford2019language}. Similarly, when applying probe sampling to AutoDAN, we use Llama2-7b-chat as the target model and GPT-2 as the draft model. For AutoPrompt, we follow their original setting, which uses RoBERTa-large~\citep{liu2019roberta} as the target model and tests on SST-2~\citep{socher2013recursive} for sentiment analysis and SICK-E~\citep{marelli-etal-2014-sick} for natural language inference. We use the RoBERTa-base model as the draft model. In the case of APE, we conduct the experiments on three widely used benchmarks GSM8K~\citep{cobbe2021gsm8k}, BBH~\citep{suzgun2023challenging}, and MMLU~\citep{hendrycks2020measuring}. For these experiments, we use Vicuna-7b-v1.5 as the target model and GPT-2 as the draft model.  

\paragraph{Evaluation.}

Following \citep{zou2023universal}, we use Attack Success Rate (ASR) as the evaluation metric for GCG and AutoDAN, which is defined as the percentage of inputs that successfully lead LLMs to generate harmful outputs. An output is determined to be harmful if it does not match with rejection phrases, such as ``I'm sorry'', ``I apologize'' and ``As an''. This is not a perfect measurement but works relatively well in practice since LLMs are trained to reject harmful replies. It is also easy to measure and interpret. For prompt learning methods tested on other tasks, we employ Accuracy (Acc) as the metric.
The processing time is determined as the average time used for each iteration across all input samples and all iterations. In all experiments, we use 1 NVIDIA A100 GPU with 80GB memory unless mentioned otherwise.

\paragraph{Hyperparameters.}

To determine the hyperparameters for probe sampling, including probe set size $k$, filtered set size reduction hyperparameter $R$, we construct a validation set of size $100$ from AdvBench by random sampling in the $400$ instances different from the test set. We follow ~\citep{zou2023universal} for the hyperparameters used in the original algorithm such as the size of the candidate set $B$. We provide detailed analysis of hyperparameters in Section \ref{sec:hyper}. When we combine probe sampling with simulated annealing, we follow the same procedure to select hyperparameters. We use the same number of optimization steps $500$ as in GCG throughout the paper. 

\subsection{Main Results}\label{sec:main_result}

\begin{table}[t]
\caption{Comparing the ASR and processing time of \texttt{Probe sampling} with and without simulated annealing to GCG with and without simulated annealing, while measuring time and FLOPs by averaging each iteration.}
  \centering
\footnotesize
\setlength{\tabcolsep}{3.8pt}
  \scalebox{0.82}{
  \begin{tabular}{ll|ccr|c|cc|cr}
    \toprule
    \multirow{3}*{\textbf{\normalsize{Model}}} &\multirow{3}*{\textbf{\normalsize{Method}}} & \multicolumn{3}{c}{\textbf{\normalsize{Harmful Strings}}} \vline & \multicolumn{5}{c}{\textbf{\normalsize{Harmful Behaviors}}} \\
    & &  & & & \textbf{Individual} & \multicolumn{2}{c}{\textbf{Multiple}} \vline & \multirow{2}*{\textbf{\normalsize{Time (s)}}}  & \multirow{2}*{\textbf{\normalsize{\#FLOPs}}}   \\
 & & ASR & \textbf{Time (s)}  & \textbf{\#FLOPs} & ASR  & ASR (train)  & ASR (test) &  &   \\
    \midrule
  \multirow{4}*{\begin{tabular}[c]{@{}l@{}} \normalsize{Vicuna}\\(7b-v1.3)
\end{tabular}}  & GCG  & $88.0$ & $4.1$ & $97.3$ T & $99.0$  & $\mathbf{100.0}$ & $98.0$ & $4.8$ & $106.8$ T\\
& GCG + Annealing  & $89.0$ & $1.5\;({2.7}\times)$ & $38.5$ T & $98.0$  & $92.0$ & $94.0$ & $2.1\;({2.3}\times)$ & $46.2$ T  \\
 & {\texttt{Probe sampling}}  & $91.0$ & $1.7\;({2.4}\times)$ & $42.4$ T & $\mathbf{100.0}$ & $96.0$ & $98.0$ & $2.3\;({2.1}\times)$ & $53.2$ T \\
 & {\texttt{PS} + Annealing} & $\mathbf{93.0}$ & $\mathbf{1.1\;({3.6}\times)}$ & $\mathbf{27.8}$ T & $\mathbf{100.0}$ & $96.0$ & $\textbf{99.0}$ & $\mathbf{1.5\;({3.2}\times)}$ & $\mathbf{24.7}$ T \\\midrule
  \multirow{4}*{\begin{tabular}[c]{@{}l@{}} \normalsize{Llama2} \\ (7b-Chat)
\end{tabular}}  & GCG & $57.0$ & $8.9$ & $198.4$ T & $69.0$ & $88.0$ & $84.0$  & $9.2$ & $202.3$ T \\
& GCG + Annealing & $55.0$ & $2.4\;({3.9}\times)$ & $39.7$ T & $68.0$ & $92.0$ & $88.0$ & $2.7\;({3.4}\times)$ & $50.6$ T \\
 & {\texttt{Probe sampling}} & $\mathbf{69.0}$ & $2.2\;({4.1}\times)$ & $43.8$ T & $\mathbf{81.0}$ & $92.0$ & $\mathbf{93.0}$ & $2.6\;({3.5}\times)$ & $40.7$ T \\
  & {\texttt{PS} + Annealing} & $64.0$ & $\mathbf{1.4\;({6.3}\times)}$ & $\mathbf{31.2}$ T & $74.0$ & $\mathbf{96.0}$ & $91.0$ & $\mathbf{1.6\;({5.6}\times)}$  & $\mathbf{32.3}$ T \\
    \bottomrule  \end{tabular}}
  \label{table:result_main}
\end{table}

\begin{table}[t]
  \centering
   \begin{minipage}[t]{0.48\textwidth}
    \centering
\caption{Transferability of \texttt{Probe sampling} with different draft models.}
  \footnotesize
  \setlength{\tabcolsep}{3.8pt}
  \scalebox{0.9}{
\begin{tabular}{l|c|cc}
\toprule
\multirow{2}*{\textbf{\normalsize{Method}}}  &  \textbf{\normalsize{Direct}} & \multicolumn{2}{c}{\textbf{\normalsize{Transfer}}}    \\
  &  {Llama2-7b} & {Vicuna-7b} & {Mistral-7b}  \\\midrule
GCG & $69.0$ & $89.0$ & $\mathbf{86.0}$ \\
\texttt{PS} (GPT-2) & $85.0$ & $92.0$ & $83.0$ \\
\texttt{PS} (ShearedLlaMa) & $\mathbf{91.0}$ & $\mathbf{93.0}$ & $85.0$ \\
\texttt{PS} (Flan-T5) & $57.0$ & $78.0$ & $69.0$  \\
\bottomrule
\end{tabular}
}\hfill
    \label{table:trans_draft}
  \end{minipage} 
\hfill
  \begin{minipage}[t]{0.48\textwidth}
    \centering
\caption{Transferability of \texttt{Probe sampling} with different filtered set size $(1-\alpha)*B/R$.}
  \footnotesize
\setlength{\tabcolsep}{3.8pt}
\scalebox{0.9}{
\begin{tabular}{l|c|cc}
\toprule
\multirow{2}*{\textbf{\normalsize{Method}}}  &  \textbf{\normalsize{Direct}} & \multicolumn{2}{c}{\textbf{\normalsize{Transfer}}}    \\
&  {Llama2-7b} & {Vicuna-7b} & {Mistral-7b} \\\midrule
GCG & $69.0$ & $89.0$ & $\mathbf{86.0}$ \\
\texttt{PS} ($R=64$) & $60.0$ & $77.0$ & $74.0$ \\
\texttt{PS} ($R=8$) & $\mathbf{85.0}$ & $\mathbf{92.0}$ & $83.0$ \\
\texttt{PS} ($R=1$) & $79.0$ & $88.0$ & $84.0$  \\
\bottomrule
\end{tabular}
}
    \label{table:trans_set}
  \end{minipage}%
\end{table}

\begin{table}[t]
  \centering
   \begin{minipage}[t]{0.42\textwidth}
    \centering
\caption{Performance of \texttt{Probe sampling} on accelerating AutoDAN.}
  \footnotesize
  \scalebox{0.82}{
\begin{tabular}{l|cc}
\toprule
{\textbf{\normalsize{Method}}}  & \multicolumn{1}{c}{\textbf{\normalsize{ASR}}} & \multicolumn{1}{c}{\textbf{\normalsize{Time (s)}}}  \\\midrule
AutoDAN-GA & $\mathbf{56.2}$ & $424.2$ \\
AutoDAN-GA + \texttt{PS} & $55.9$ & $\mathbf{182.7\;(2.3\times)}$ \\\midrule
AutoDAN-HGA & $60.8$ & $237.9$ \\
AutoDAN-HGA + \texttt{PS} & $\mathbf{62.1}$ & $\mathbf{95.3\;(2.5\times)}$ \\
\bottomrule
\end{tabular}
}
    \label{table:autodan}
  \end{minipage} 
\hfill
  \begin{minipage}[t]{0.56\textwidth}
    \centering
\caption{Performance of \texttt{Probe sampling} on accelerating prompt learning method AutoPrompt.}
  \footnotesize
  \scalebox{0.82}{
\begin{tabular}{l|cc|cc}
\toprule
\multirow{2}*{\textbf{\normalsize{Method}}} & \multicolumn{2}{c}{\textbf{\normalsize{SST-2}}} & \multicolumn{2}{c}{\textbf{\normalsize{SICK-E}}} \\
 & \textbf{Acc} & \textbf{Time (s)} & \textbf{Acc} & \textbf{Time (s)} \\\midrule
Original & $85.2$ & N / A & $49.4$ & N / A \\\midrule
Autoprompt & $\mathbf{91.4}$ & $228.4$ & $\mathbf{69.3}$ & $42.7$ \\
Autoprompt + \texttt{PS} & $90.6$ & $\mathbf{127.2\;(1.8\times)}$ & $68.9$ & $\mathbf{23.6\;(1.8\times)}$ \\ \bottomrule
\end{tabular}
}
    \label{table:autoprompt}
  \end{minipage}%
\end{table}

\begin{table}[t]
  \centering
  \caption{Performance of \texttt{Probe sampling} on accelerating prompt learning method APE. }
  \footnotesize
  \scalebox{0.9}{
  \begin{tabular}{l|cc|cc|cc}
    \toprule
     \multirow{2}*{\textbf{\normalsize{Method}}} & \multicolumn{2}{c}{\textbf{\normalsize{GSM8K}}} \vline & \multicolumn{2}{c}{\textbf{\normalsize{MMLU}}} \vline & \multicolumn{2}{c}{\textbf{\normalsize{BBH}}} \\
 & \textbf{Acc} & \textbf{Time (s)} & \textbf{Acc} & \textbf{Time (s)} & \textbf{Acc} & \textbf{Time (s)}  \\\midrule
 Vicuna & $20.4$ & N/A & $45.6$ & N / A & $38.6$ & N / A \\
 APE  & $21.3$ & $431.8$ & $\mathbf{48.2}$ & $187.3$ & $\mathbf{40.8}$ & $265.2$ \\
 APE+\texttt{PS} & $\mathbf{22.4}$ & $\mathbf{192.3\;(2.3\times)}$ & $47.3$ & $\mathbf{102.5\;(1.8\times)}$ & $39.9$ & $\mathbf{88.7\;(3.0\times)}$  \\\bottomrule  
 \end{tabular}
  }
  \label{table:ape}
\end{table}

\paragraph{Acceleration results.}
As shown in Table \ref{table:result_main}, probe sampling achieves a speedup of $5.6$ times and $6.3$ times on Human Behaviors and Human Strings with Llama2 when combined with simulated annealing. Probe sampling achieves a speedup of $3.5$ and $4.1$ times alone. With Vicuna, we achieve an overall speedup of $3.2$ and $3.6$ respectively on the two datasets. We also measure the \#FLOPs for different settings and found that the speedup results reflects in the reduction of \#FLOPs. For example, with Llama2, the \#FLOPs reduction is $202.3\mathrm{T} / 32.3 \mathrm{T} = 6.3$ times and $198.4 \mathrm{T} / 31.2 \mathrm{T} = 6.4$ times on the two sets, which is close to the actual speedup results. This also shows that our algorithm results in little overhead with the introduced new procedures. 
It is worth noting that simulated annealing also achieves decent acceleration and is complementary to our acceleration results.

\paragraph{GCG results. } Interestingly, we achieve a better ASR score than the GCG algorithm although technically acceleration introduces noise to the algorithm. For instance, with Llama2, we improve the ASR from $57.0$ to $64.0$ on Human Strings and from $84.0$ to $91.0$ on Human Behaviors. We hypothesize that the improvement comes from the randomness added to the GCG algorithm based on greedy search over a single objective. Introducing randomness and noise has been seen as one of the advantages of SGD over full batch training. In contrast, simulated annealing only leads to comparable ASR when applied on GCG. 

\paragraph{Transferability} Table \ref{table:trans_draft} shows probe sampling's transferability across draft models based on Llama2-7b-Chat to various target models. We find that it maintains transferability when using draft models like GPT-2 and SheardLlaMa, which preserve the original ASR of plain GCG. However, draft models that significantly degrade initial performance, such as Flan-T5, impair transferability. Table \ref{table:trans_set} examines probe sampling transferability across filtered set sizes. Results align with prior findings: probe sampling minimally impacts transferability with appropriate parameters but decreases performance when Llama2-7b-chat's direct ASR is low such as $R=64$.

\paragraph{Results on AutoDAN, Autoprompt and APE.} Table \ref{table:autoprompt} demonstrates the effective acceleration of AutoPrompt through the implementation of probe sampling, resulting in a speedup of $1.79\times$ on SST-2 and $1.83\times$ on SICK-E. Importantly, this acceleration is achieved without compromising performance, as evidenced by the minimal changes in accuracy from $91.4$ to $90.6$ on SST-2 and from $69.3$ to $68.9$ on SICK-E. Furthermore, the application of probe sampling to APE, as presented in Table \ref{table:ape}, results in significant speed improvements, with a speedup of $2.3\times$ on GSM8K, $1.8\times$ on MMLU, and $3.0\times$ on BBH. Similarly, these speed enhancements do not compromise the performance of APE. In addition, we implement probe sampling on another jailbreak method, AutoDAN. The detailed results can be found in Table \ref{table:autodan}. Our findings indicate that probe sampling can achieve a speedup of $2.3\times$ for AutoDAN-GA and $2.5\times$ for AutoDAN-HGA, while minimally affecting its performance. These results demonstrate the effectiveness of our method in not only accelerating GCG but also its applicability to general prompt optimization methods and other LLM attack methods.

\subsection{Computation Detail Analysis}
\begin{figure}[t!]
\hspace*{-0.8cm}
    \centering
    \begin{subfigure}[t]{0.48\textwidth}
    \begin{tikzpicture}
    \pgfplotsset{width=6.8cm,height=4cm}
    \begin{axis}[
        xbar stacked, 
        bar width = 15pt,
        ymin=0.5, ymax=3.5,
        yticklabels={PS + Annealing, Probe sampling, GCG},
        ytick=data,
        xmin=0, xmax=100,
        xticklabel style = {font=\fontsize{7.5}{1}\selectfont},
        yticklabel style = {font=\fontsize{6.5}{1}\selectfont},
        legend style={font=\fontsize{9}{1}\selectfont},
        legend style={at={(1.2, 1.3)},
          anchor=north,legend columns=3},
        legend cell align={left},
        xticklabel={$\pgfmathprintnumber{\tick}\%$},
        nodes near coords={\pgfmathprintnumber\pgfplotspointmeta\%}, 
        every node near coord/.append style={font=\fontsize{6}{1}\selectfont},
      ]
      \addplot[fill=dm1]  coordinates {(43, 1) (38, 2) (66, 3)};
      \addplot[fill=dm2]  coordinates {(8, 1) (25, 2) (0, 3)};
      \addplot[fill=dm3]  coordinates {(49, 1) (37, 2) (34, 3)};
    \legend{Target Model, Draft Model, Vacant}
    \end{axis}
    \end{tikzpicture}
    \caption{Llama2-7b-chat}
    \label{fig:memory_chat1}
    \end{subfigure}
~
    \begin{subfigure}[t]{0.48\textwidth}
    \begin{tikzpicture}
    \pgfplotsset{width=6.8cm,height=4cm}
    \begin{axis}[
        xbar stacked, 
        bar width = 15pt,
        ymin=0.5, ymax=3.5,
        yticklabels={PS + Annealing, Probe sampling, GCG},
        ytick=data,
        xmin=0, xmax=100,
        xticklabel style = {font=\fontsize{7.5}{1}\selectfont},
        yticklabel style = {font=\fontsize{6.5}{1}\selectfont},
        xticklabel={$\pgfmathprintnumber{\tick}\%$},
        nodes near coords={\pgfmathprintnumber\pgfplotspointmeta\%}, 
        every node near coord/.append style={font=\fontsize{6}{1}\selectfont},
      ]
      \addplot[fill=dm1]  coordinates {(36, 1) (36, 2) (48, 3)};
      \addplot[fill=dm2]  coordinates {(5, 1) (15, 2) (0, 3)};
      \addplot[fill=dm3]  coordinates {(59, 1) (49, 2) (52, 3)};
   
    \end{axis}
    \end{tikzpicture}
    \caption{Vicuna-7b-v1.3}
    \label{fig:memory_vicuna1}
    \end{subfigure}
\caption{Memory usage on a single A100 with 80GB memory with (a) Llama2-7b-chat and (b) Vicuna-7b-v1.3 on $1$ instance. The memory consumption of probe sampling with or without simulated annealing is similar to that of the original setting. The computation with the target model still takes most of the memory.
}
    \label{fig:memory}
\end{figure}
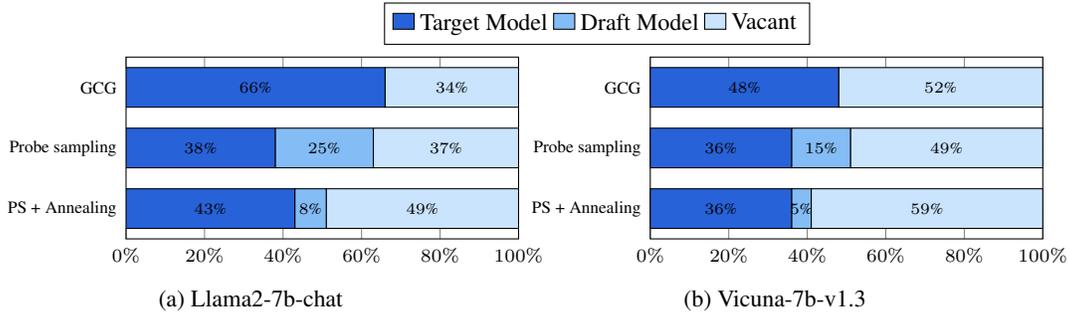

\paragraph{Memory allocation.}
We evaluate whether probe sampling uses more memory because of the use of an extra model. In Figure \ref{fig:memory}, we show the memory usage of GCG, probe sampling with and without annealing using either {Llama2-7b-chat} and {Vicuna-7b-v1.3}. Probe sampling uses a similar amount of memory to the original GCG algorithm although it involves extra procedures and an extra model, by saving the computation of target model on the whole candidate set. 
As such, the usage of probe sampling does not introduce extra memory and can be applied when the original GCG algorithm is applied. 
In terms of the memory usage of the target model and the draft model, most of the memory is spent on target model, probably because the draft model is much smaller.

\begin{figure*}[t]
\centering
\includegraphics[width=0.85\textwidth]{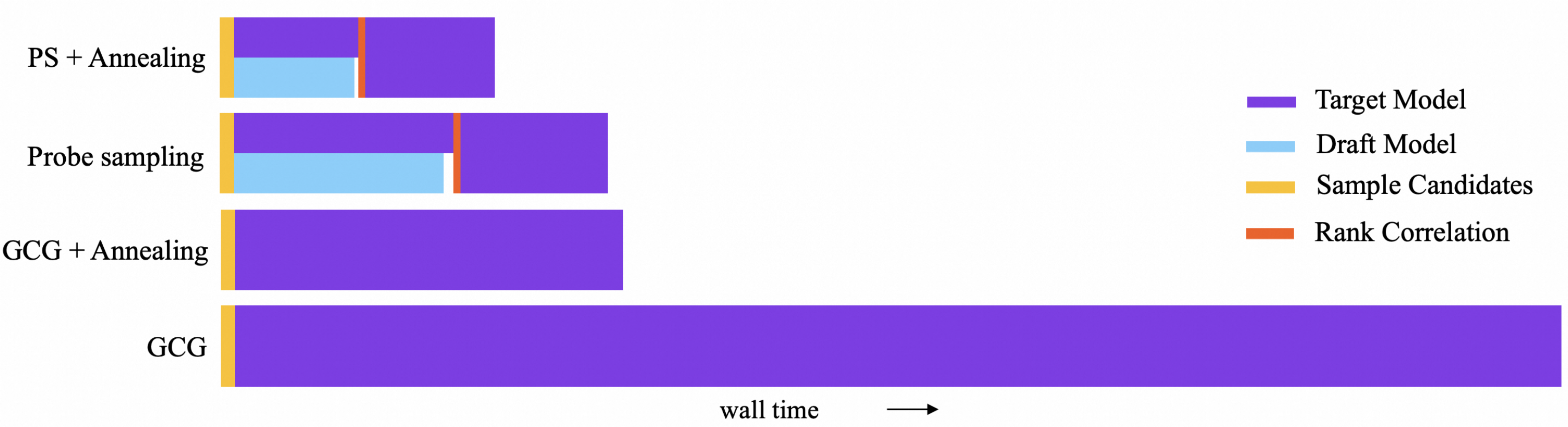}
  \caption{Wall time of GCG, probe sampling with and without simulated annealing. For the target model computation, the first part is done on the probe set and the second part is done on the filtered set. Draft model computation and computation of the target model on the probe set are suited to be done in parallel as they take similar time. } 
\label{fig:time_chat}
\end{figure*}

\paragraph{Time allocation.}
We look at the specific time spent on different operations. As shown in Figure \ref{fig:time_chat}, probe set computation using the target model and full set computation using the draft model take a similar amount of time so we can parallelize the computation easily. Sampling candidates in the graph involves a forward and backward pass as mentioned earlier and can be completed relatively quickly. Similarly, it is also fast to compute the agreement using the ranked losses on CPU, so our algorithm introduces relatively little overhead.

\subsection{Further analysis}\label{sec:hyper}
In this section, we conduct extensive studies to understand how the proposed method works. We conduct all of the following experiments on the validation set, so the numbers are not directly comparable to the numbers in the main results. For the validation set, the original GCG algorithm achieves an ASR of $66.0$ with an average time of $9.16$ seconds per iteration. 
In each of the study, we highlight the settings that we find to be the best. 

\paragraph{Filtered set size.}
The filtered set size is the most important factor in our method. If it is too small, then we will achieve a lot of speedup at the cost of relying too heavily on the draft model and resulting in a lower ASR. If it is too big, then we would not achieve much speedup. Hence we experiment with different filtered size reduction hyperparameter $R$. The filter set size is $(1-\alpha) * B / R$ where $\alpha$ is the probe agreement score described in Section \ref{sec:probe_sampling}. 

As shown in Table \ref{table:filtered_size}, the time does monotonically decrease if we use a smaller filtered set size. However, interestingly, there is a sweetspot for the ASR with $R$ set to $8$. We believe that this can resonates with the hypothesis of introducing randomness as the source of ASR boosts. Both too much or too little randomness hurt performance. As such, we use $R=8$ for probe sampling. We further show several convergence processes with varying values of $R$ in Appendix \ref{sec:appen_converge}.

\begin{table}[t]
  \centering
  
  \begin{minipage}[t]{0.49\textwidth}
    \centering
    \caption{Ablation on the filtered set size reduction $R$. The filter set size is $(1-\alpha) * B / R$.}
    \footnotesize
    \scalebox{0.8}{
      \begin{tabular}{l|c|c|c|c|c|c}
        \toprule
        \textbf{Reduction $R$} & $64$ & $16$ & \cellcolor{gray!50} $8$  & $4$ & $2$ & $1$ \\
        \midrule
        ASR & $60.0$ & $70.0$ & \cellcolor{gray!50} $\mathbf{85.0}$ & $81.0$ & $76.0$ & $79.0$ \\
        \midrule
        Time (s) & $\mathbf{2.01}$ & $2.31$ & \cellcolor{gray!50} $2.60$ & $3.02$ & $3.41$ & $5.19$ \\
        \bottomrule
      \end{tabular}
    }
    \label{table:filtered_size}
  \end{minipage}%
  \hfill 
  \begin{minipage}[t]{0.49\textwidth}
    \centering
    \caption{Ablation on fixed probe agreement score $\alpha$ vs adaptive score.}
    \footnotesize
    \scalebox{0.8}{
      \begin{tabular}{l|c|c|c|c|c}
        \toprule
        \textbf{Agreement $\alpha$} & $0.9$ & $0.6$ & $0.3$  & $0.0$ & \cellcolor{gray!50} Adaptive \\
        \midrule
        ASR & $70.0$ & $77.0$ & $75.0$ & $81.0$ & \cellcolor{gray!50} $\mathbf{85.0}$ \\
        \midrule
        Time (s) & $\mathbf{2.17}$ & $2.41$ & $2.71$ & $3.01$ & \cellcolor{gray!50} $2.60$ \\
        \bottomrule
      \end{tabular}
    }
    \label{table:adaptive}
  \end{minipage}
\end{table}

\paragraph{Adaptive vs fixed filtered set size.}
As mentioned in Section \ref{sec:probe_sampling}, an alternative to use an adaptive filtered set size is to use a fixed size. Here we investigate whether it matters to use an adaptive filtered set size that is determined by how much the draft model and the target model agree on each candidate set. 
To use a fixed size, we simply fix the probe agreement score $\alpha$ to be $0.9$, $0.6$, $0.3$, and $0.0$ and compare with the adaptive case. As shown in Table \ref{table:adaptive}, fixed probe agreement scores always lead to worse ASR. Furthermore, when adopting GPT-2 as the draft model, the average agreement score is $0.45$ with a standard deviation of $0.11$. This shows that the agreement score between the two models varies significantly for different candidate sets. We also provide the statistics of $\alpha$ for other draft models in Table \ref{table:abla_draft}.

\paragraph{Probe agreement measurement. }
We also experiment alternatives to measure the probe agreement score, including the Pearson correlation coefficient~\citep{pearson1900x}, 
Kendall's Tau correlation coefficient~\citep{kendall1938new}, and Goodman and Kruskal's gamma~\citep{goodman1979measures} where the Pearson correlation coefficient directly uses the loss values to compute the agreement and the others use the ranking information. As shown in Table \ref{table:abla_cor}, all methods have similar time cost, and Spearman's rank correlation coefficient achieves the best ASR. The Pearson correlation coefficient performs worse than other ranking-based agreement measurement. 

\paragraph{Probe set size.}
The size of the probe set also determines whether the probe agreement score is measured accurately. As such, we experiment with different probe set size and report the performance in Table \ref{table:abla_probe}. 
We find that using a small probe set such as $B/64$ or $B/32$ can result in inaccurate agreement score, which put a put a significant toll on the attack success rate. It also does not lead to too much time reduction since the draft model computation done in parallel takes more time and the reduced computation is not the bottleneck. 
Using a larger probe set size such as $B/4$ and $B/2$ will lead to more accurate agreement score but does not increase the ASR significantly. As such, using a probe set of size $B/16$ is good enough to accurately measure the agreement and achieves maximum time reduction.

\paragraph{Draft model study.}
Here we also experiment with bigger draft models, some of which is of similar size to Llama2. We experiment with GPT-Neo~\citep{gao2020pile}, Flan-T5-base~\citep{chung2022scaling}, BART~\citep{lewis2019bart}, Phi-1.5~\citep{textbooks2}, TinyLlama~\citep{zhang2024tinyllama} and Sheared-LLaMA~\citep{xia2023sheared}. Among them, Sheared-LLaMA might be the closest to Llama2 since it is a pruned version of Llama2. For TinyLlama, Phi and Sheared-LLaMA, we use 2 A100s with 80GB memory to fit the whole computation. 

As shown in Table \ref{table:abla_draft}, Sheared-LlaMa achieves the best ASR although the time reduction is not as good as smaller models such as GPT-2 and there would be a higher time cost if we manage to fit all computation in one GPU. 
On contrast, Flan-T5, BART, TinyLlama and Mistral all achieve lower ASRs probably because of being very different than Llama2. However, the results are still better than the baseline ASR $66.0$. GPT-2 and GPT-Neo achieve a good balance of performance and speedup. 



\begin{table}[t]
  \centering  

  \begin{minipage}[t]{0.48\textwidth}
    \centering
    \caption{Ablation on probe agreement measurements. All methods achieve similar speedup while Spearman's rank correlation coefficient achieves the best ASR.}
    \footnotesize
    \scalebox{0.82}{
      \begin{tabular}{l|c|c|c|c}
        \toprule
        \textbf{Cor} & \cellcolor{gray!50} Spearman & Pearson & Kendall & Kruskal \\
        \midrule
        ASR & \cellcolor{gray!50} $\mathbf{85.0}$ & $70.0$ & $74.0$ & $79.0$ \\
        \midrule
        Time (s) & \cellcolor{gray!50} $2.60$ & $2.47$ & $2.53$ & $\mathbf{2.43}$ \\
        \bottomrule
      \end{tabular}
    }
    \label{table:abla_cor}
  \end{minipage}
  \hfill
  \begin{minipage}[t]{0.48\textwidth}
    \centering
    \caption{Ablation on the probe set size $k$. Using $B/16$ leads to accurate probe agreement measurement while achieving significant acceleration.}
    \footnotesize
    \scalebox{0.82}{
      \begin{tabular}{l|c|c|c|c|c|c}
        \toprule
        \textbf{Probe} & $B/64$ & $B/32$ & \cellcolor{gray!50} $B/16$ & $B/4$ & $B/2$ & $B$ \\
        \midrule
        ASR & $64.0$ & $72.0$ & \cellcolor{gray!50} $85.0$ & $86.0$ & $85.0$ & $\mathbf{87.0}$ \\
        \midrule
        Time (s) & $\mathbf{2.10}$ & $2.57$ & \cellcolor{gray!50} $2.60$ & $3.41$ & $5.61$ & $9.58$ \\
        \bottomrule
      \end{tabular}
    }
    \label{table:abla_probe}
  \end{minipage}
\end{table}

\begin{table}[t]
  \centering
  \caption{Experiments with different draft models. Models with over $1$B parameters, like TinyLlama, Phi, and ShearedLlMa, need two GPUs for parallel computation. ShearedLlMa achieves the highest ASR probably because it is a pruned version of Llama2. Both GPT-2 and GPT-Neo achieve a good balance of ASR and speedup. }
  \footnotesize
  \scalebox{0.84}{
    \begin{tabular}{l|c|c|c|c|c|c|c}
      \toprule
       & \multicolumn{4}{c}{\textbf{1 GPU}} \vline & \multicolumn{3}{c}{\textbf{2 GPUs}} \\
     \textbf{Model} & \cellcolor{gray!50} \begin{tabular}{cc} GPT-2  \\ ($124$M)
\end{tabular} & \begin{tabular}{cc} GPT-Neo  \\ ($125$M)
\end{tabular}  & \begin{tabular}{cc} Flan-T5  \\ ($248$M)
\end{tabular} & \begin{tabular}{cc} BART\\ ($406$M)
\end{tabular}    & \begin{tabular}{cc} TinyLlama\\ ($1.1$B)
\end{tabular}  & \begin{tabular}{cc} Phi\\ ($1.3$B)
\end{tabular} &\begin{tabular}[c]{cc} ShearedLlaMa\\ ($1.3$B)
\end{tabular}  \\
      \midrule
        $\alpha$ & \cellcolor{gray!50} $0.45\pm0.10$ & $0.51\pm0.11$ & $0.61\pm0.13$ & $0.46\pm0.09$ & $0.52\pm0.13$ & $0.52\pm0.11$ & $\textbf{0.35}\pm0.12$  \\\midrule
      ASR & \cellcolor{gray!50} $85.0$ & $81.0$ & $57.0$ & $76.0$ & $72.0$ & $82.0$ & $\mathbf{91.0}$  \\\midrule
      Time (s) & \cellcolor{gray!50} $\mathbf{2.60}$ & $2.82$ & $3.89$ & $2.93$  & $3.38$ & $4.83$ & $3.93$   \\
      \bottomrule
    \end{tabular}
  }
  \vspace{-0.2cm}
  \label{table:abla_draft}
\end{table}

\section{Related Work}

\paragraph{Alignment of LLMs.}
To build safe LLMs, alignments has also been a widely studied topic in the community~\citep{stiennon2020learning, ouyang2022training}. Efforts have been put into improving helpfulness~\citep{bai2022training, cheng2023adversarial}, honesty~\citep{kaddour2023challenges, liu2023trustworthy, xu2023align}, and harmlessness~\citep{hartvigsen2022toxigen}. 
Among these works, there has been a growing interest in using feedback from a LLM to perform alignment~\citep{bai2022constitutional, gulcehre2023reinforced, burns2023weak, yuan2024self}. 
Despite all the efforts, there has not been a definitive answer for LLM safety alignments, which also motivates our research in LLM safety. 

\paragraph{Discrete Prompt Optimization.} Attacking LLMs via adversarial prompt can be formulated as a discrete prompt optimization problem~\citep{zou2023universal}. In this context, attacking algorithms strive to discover superior prompts that effectively steer aligned LLMs toward generating adversarial answers. Some approaches leverage LLMs themselves to iteratively refine prompts~\citep{xu2022gps, pryzant2023automatic}. However, aligned LLMs may resist refining adversarial prompts, rendering these methods ineffective. Other strategies employ RL-based prompt optimization techniques such as those in~\citep{mingkai2022rlprompt, ludynamic}, necessitating additional MLP training with extensive adversarial data and specific reward design. Moreover, other models introduced in~\citep{cho2023discrete, do2023prompt} to help with prompt optimization must remain unaligned, particularly in jailbreak scenarios~\citep{chao2023jailbreaking}. However, their performance tends to be limited, especially when dealing with strongly fine-tuned models like Llama2-Chat.

\paragraph{LLM Jailbreaks.} LLM Jailbreaks have received considerable interests recently since due to the implications of applying LLMs widely in human society. Although there is a continuous effort to build safe and reliable LLMs,  bypassing the safety mechanism of LLMs is not uncommon. For example, fine-tuning a safe LLM on a few data instances can easily breaks its safety guarantees~\citep{qi2023fine, lermen2023lora}. Treating the jailbreak as a prompt optimization problem has also led to a certain level of success~\citep{zou2023universal, mokander2023auditing, liu2023autodan, chao2023jailbreaking, geisler2024attacking}. In addition, conversing in a ciphered language~\citep{yuan2023gpt}, planting a backdoor during RLHF~\citep{rando2023universal}, using a less well-aligned language~\citep{deng2023multilingual} and multi-modality~\citep{shayegani2023jailbreak} can also lead to successful jailbreaks. Researchers also construct large dataset of manual jailbreak prompts~\citep{toyer2023tensor}.

Among these jailbreak methods, the prompt optimization method GCG~\citep{zou2023universal} provides the more general and universal solution for us to study the jailbreaking problem. As such, in this work, we mainly focus on the acceleration of GCG, but the idea of delegating computation to a draft model can also be applied in other situations such as the multi-modality case and finetuning case. We leave the extension of this work for future work.

\paragraph{Acceleration.}
In the field of acceleration, speculative sampling~\citep{chen2023accelerating, leviathan2023fast} is the most relevant to our method. They also use a draft model but its design cannot be directly applied to accelerate the GCG algorithm. REST~\citep{he2023rest} adopts the concept of speculative sampling but uses a retrieval approach based on a Trie to construct the candidate. The attention module has also been a focus of acceleration because of its quadratic nature~\citep{dao2022flashattention, cai2024medusa}. There have also been continuous interests in more efficient versions of Transformers~\citep{so2019evolved, dai2021coatnet, liu2021pay, gu2020hippo, gu2021efficiently}. These architectural changes are complementary to our algorithm design and we leave it to future work.

\section{Conclusion}\label{sec:discussion}
In this paper, we propose an algorithm probe sampling that can effectively accelerate the GCG algorithm. We achieve an acceleration ranging from $2.1 \times$ to $6.3 \times$ in different scenarios on AdvBench. We illustrate the intuition and how the algorithm works through extensive experiments. Furthermore, this approach is also applied to general prompt optimization methods and other jailbreak techniques, including AutoPrompt, APE, and AutoDAN.
We believe the idea of using the probe agreement score to perform adaptive computation can be applied to cases other than GCG. For example, it could potentially be used to perform conditional computation for attention. Another direction is to extend the framework to the multi-modality case which can be interesting given the vast amount of video data. It would also be interesting to run a small draft model on the scale of web data to detect the existence of natural adversarial prompts.

\section*{Limitation and Impact Statements}
Probe sampling has two main limitations. Firstly, it exhibits relatively slow performance when tested on large-sized test sets, which hampers its efficiency. Secondly, it is limited to supporting only open-source models, thereby excluding proprietary or closed-source models from benefiting from the proposed acceleration techniques. These limitations indicate the need for further improvements to enhance the speed and broaden the model support in order to make the jailbreak acceleration approach more robust and applicable across a wider range of language models. 

Probe sampling can be applied to accelerate GCG algorithm. Having a faster algorithm to explore adversarial cases of alignments enable us to study how to make LLMs safer. As far as we know, as of now, there is not a LLM that can use this algorithm to achieve malicious behavior in real-world that would not be possible without the algorithm. The goal of this research is to present a general algorithm which may inspire new research, and also contribute to the gradual progress of building safe and aligned AIs.

\clearpage

\section*{Acknowledgements}
This research is partially supported by the National Research Foundation Singapore under the AI Singapore Programme (AISG Award No: AISG2-TC-2023-010-SGIL) and the Singapore Ministry of Education Academic Research Fund Tier 1 (Award No: T1 251RES2207). Xuan Long Do is supported by the A*STAR Computing and Information Science (ACIS) scholarship. We thank Liwei Kang for insightful discussion, Liying Cheng for helping with plotting figures.

\bibliography{main}

\begin{thebibliography}{71}
\expandafter\ifx\csname natexlab\endcsname\relax\def\natexlab#1{#1}\fi

\bibitem[{Bai et~al.(2022{\natexlab{a}})Bai, Jones, Ndousse, Askell, Chen,
  DasSarma, Drain, Fort, Ganguli, Henighan et~al.}]{bai2022training}
Yuntao Bai, Andy Jones, Kamal Ndousse, Amanda Askell, Anna Chen, Nova DasSarma,
  Dawn Drain, Stanislav Fort, Deep Ganguli, Tom Henighan, et~al.
  2022{\natexlab{a}}.
\newblock Training a helpful and harmless assistant with reinforcement learning
  from human feedback.
\newblock \emph{arXiv preprint arXiv:2204.05862}.

\bibitem[{Bai et~al.(2022{\natexlab{b}})Bai, Kadavath, Kundu, Askell, Kernion,
  Jones, Chen, Goldie, Mirhoseini, McKinnon et~al.}]{bai2022constitutional}
Yuntao Bai, Saurav Kadavath, Sandipan Kundu, Amanda Askell, Jackson Kernion,
  Andy Jones, Anna Chen, Anna Goldie, Azalia Mirhoseini, Cameron McKinnon,
  et~al. 2022{\natexlab{b}}.
\newblock Constitutional ai: Harmlessness from ai feedback.
\newblock \emph{arXiv preprint arXiv:2212.08073}.

\bibitem[{Brown et~al.(2020)Brown, Mann, Ryder, Subbiah, Kaplan, Dhariwal,
  Neelakantan, Shyam, Sastry, Askell et~al.}]{brown2020language}
Tom Brown, Benjamin Mann, Nick Ryder, Melanie Subbiah, Jared~D Kaplan, Prafulla
  Dhariwal, Arvind Neelakantan, Pranav Shyam, Girish Sastry, Amanda Askell,
  et~al. 2020.
\newblock Language models are few-shot learners.
\newblock \emph{Advances in neural information processing systems},
  33:1877--1901.

\bibitem[{Burns et~al.(2024)Burns, Izmailov, Kirchner, Baker, Gao,
  Aschenbrenner, Chen, Ecoffet, Joglekar, Leike et~al.}]{burns2023weak}
Collin Burns, Pavel Izmailov, Jan~Hendrik Kirchner, Bowen Baker, Leo Gao,
  Leopold Aschenbrenner, Yining Chen, Adrien Ecoffet, Manas Joglekar, Jan
  Leike, et~al. 2024.
\newblock Weak-to-strong generalization: Eliciting strong capabilities with
  weak supervision.
\newblock In \emph{Forty-first International Conference on Machine Learning}.

\bibitem[{Cai et~al.(2024)Cai, Li, Geng, Peng, Lee, Chen, and
  Dao}]{cai2024medusa}
Tianle Cai, Yuhong Li, Zhengyang Geng, Hongwu Peng, Jason~D Lee, Deming Chen,
  and Tri Dao. 2024.
\newblock Medusa: Simple llm inference acceleration framework with multiple
  decoding heads.
\newblock In \emph{Forty-first International Conference on Machine Learning}.

\bibitem[{Chao et~al.()Chao, Robey, Dobriban, Hassani, Pappas, and
  Wong}]{chao2023jailbreaking}
Patrick Chao, Alexander Robey, Edgar Dobriban, Hamed Hassani, George~J Pappas,
  and Eric Wong.
\newblock Jailbreaking black box large language models in twenty queries.
\newblock In \emph{R0-FoMo: Robustness of Few-shot and Zero-shot Learning in
  Large Foundation Models}.

\bibitem[{Chen et~al.(2023)Chen, Borgeaud, Irving, Lespiau, Sifre, and
  Jumper}]{chen2023accelerating}
Charlie Chen, Sebastian Borgeaud, Geoffrey Irving, Jean-Baptiste Lespiau,
  Laurent Sifre, and John Jumper. 2023.
\newblock Accelerating large language model decoding with speculative sampling.
\newblock \emph{arXiv preprint arXiv:2302.01318}.

\bibitem[{Cheng et~al.(2023)Cheng, Yang, Li, Dai, and
  Du}]{cheng2023adversarial}
Pengyu Cheng, Yifan Yang, Jian Li, Yong Dai, and Nan Du. 2023.
\newblock Adversarial preference optimization.
\newblock \emph{arXiv preprint arXiv:2311.08045}.

\bibitem[{Cho et~al.(2023)Cho, Jeong, yeon Seo, and Park}]{cho2023discrete}
Sukmin Cho, Soyeong Jeong, Jeong yeon Seo, and Jong Park. 2023.
\newblock Discrete prompt optimization via constrained generation for zero-shot
  re-ranker.
\newblock In \emph{The 61st Annual Meeting Of The Association For Computational
  Linguistics}.

\bibitem[{Chowdhery et~al.(2023)Chowdhery, Narang, Devlin, Bosma, Mishra,
  Roberts, Barham, Chung, Sutton, Gehrmann et~al.}]{chowdhery2022palm}
Aakanksha Chowdhery, Sharan Narang, Jacob Devlin, Maarten Bosma, Gaurav Mishra,
  Adam Roberts, Paul Barham, Hyung~Won Chung, Charles Sutton, Sebastian
  Gehrmann, et~al. 2023.
\newblock Palm: Scaling language modeling with pathways.
\newblock \emph{Journal of Machine Learning Research}, 24(240):1--113.

\bibitem[{Chung et~al.(2024)Chung, Hou, Longpre, Zoph, Tay, Fedus, Li, Wang,
  Dehghani, Brahma et~al.}]{chung2022scaling}
Hyung~Won Chung, Le~Hou, Shayne Longpre, Barret Zoph, Yi~Tay, William Fedus,
  Yunxuan Li, Xuezhi Wang, Mostafa Dehghani, Siddhartha Brahma, et~al. 2024.
\newblock Scaling instruction-finetuned language models.
\newblock \emph{Journal of Machine Learning Research}, 25(70):1--53.

\bibitem[{Cobbe et~al.(2021)Cobbe, Kosaraju, Bavarian, Chen, Jun, Kaiser,
  Plappert, Tworek, Hilton, Nakano, Hesse, and Schulman}]{cobbe2021gsm8k}
Karl Cobbe, Vineet Kosaraju, Mohammad Bavarian, Mark Chen, Heewoo Jun, Lukasz
  Kaiser, Matthias Plappert, Jerry Tworek, Jacob Hilton, Reiichiro Nakano,
  Christopher Hesse, and John Schulman. 2021.
\newblock Training verifiers to solve math word problems.
\newblock \emph{arXiv preprint arXiv:2110.14168}.

\bibitem[{Dai et~al.(2021)Dai, Liu, Le, and Tan}]{dai2021coatnet}
Zihang Dai, Hanxiao Liu, Quoc~V Le, and Mingxing Tan. 2021.
\newblock Coatnet: Marrying convolution and attention for all data sizes.
\newblock \emph{Advances in neural information processing systems},
  34:3965--3977.

\bibitem[{Dao et~al.(2022)Dao, Fu, Ermon, Rudra, and
  R{\'e}}]{dao2022flashattention}
Tri Dao, Dan Fu, Stefano Ermon, Atri Rudra, and Christopher R{\'e}. 2022.
\newblock Flashattention: Fast and memory-efficient exact attention with
  io-awareness.
\newblock \emph{Advances in Neural Information Processing Systems},
  35:16344--16359.

\bibitem[{Deng et~al.(2024)Deng, Zhang, Pan, and Bing}]{deng2023multilingual}
Yue Deng, Wenxuan Zhang, Sinno~Jialin Pan, and Lidong Bing. 2024.
\newblock Multilingual jailbreak challenges in large language models.
\newblock In \emph{The Twelfth International Conference on Learning
  Representations}.

\bibitem[{Gao et~al.(2020)Gao, Biderman, Black, Golding, Hoppe, Foster, Phang,
  He, Thite, Nabeshima et~al.}]{gao2020pile}
Leo Gao, Stella Biderman, Sid Black, Laurence Golding, Travis Hoppe, Charles
  Foster, Jason Phang, Horace He, Anish Thite, Noa Nabeshima, et~al. 2020.
\newblock The pile: An 800gb dataset of diverse text for language modeling.
\newblock \emph{arXiv preprint arXiv:2101.00027}.

\bibitem[{Geisler et~al.(2024)Geisler, Wollschl{\"a}ger, Abdalla, Gasteiger,
  and G{\"u}nnemann}]{geisler2024attacking}
Simon Geisler, Tom Wollschl{\"a}ger, MHI Abdalla, Johannes Gasteiger, and
  Stephan G{\"u}nnemann. 2024.
\newblock Attacking large language models with projected gradient descent.
\newblock \emph{arXiv preprint arXiv:2402.09154}.

\bibitem[{Goodman et~al.(1979)Goodman, Kruskal, Goodman, and
  Kruskal}]{goodman1979measures}
Leo~A Goodman, William~H Kruskal, Leo~A Goodman, and William~H Kruskal. 1979.
\newblock \emph{Measures of association for cross classifications}.
\newblock Springer.

\bibitem[{Gu et~al.(2020)Gu, Dao, Ermon, Rudra, and R{\'e}}]{gu2020hippo}
Albert Gu, Tri Dao, Stefano Ermon, Atri Rudra, and Christopher R{\'e}. 2020.
\newblock Hippo: Recurrent memory with optimal polynomial projections.
\newblock \emph{Advances in neural information processing systems},
  33:1474--1487.

\bibitem[{Gu et~al.(2021)Gu, Goel, and R{\'e}}]{gu2021efficiently}
Albert Gu, Karan Goel, and Christopher R{\'e}. 2021.
\newblock Efficiently modeling long sequences with structured state spaces.
\newblock \emph{arXiv preprint arXiv:2111.00396}.

\bibitem[{Gulcehre et~al.(2023)Gulcehre, Paine, Srinivasan, Konyushkova,
  Weerts, Sharma, Siddhant, Ahern, Wang, Gu et~al.}]{gulcehre2023reinforced}
Caglar Gulcehre, Tom~Le Paine, Srivatsan Srinivasan, Ksenia Konyushkova, Lotte
  Weerts, Abhishek Sharma, Aditya Siddhant, Alex Ahern, Miaosen Wang, Chenjie
  Gu, et~al. 2023.
\newblock Reinforced self-training (rest) for language modeling.
\newblock \emph{arXiv preprint arXiv:2308.08998}.

\bibitem[{Guo et~al.(2021)Guo, Sablayrolles, J{\'e}gou, and
  Kiela}]{guo2021gradient}
Chuan Guo, Alexandre Sablayrolles, Herv{\'e} J{\'e}gou, and Douwe Kiela. 2021.
\newblock Gradient-based adversarial attacks against text transformers.
\newblock In \emph{Proceedings of the 2021 Conference on Empirical Methods in
  Natural Language Processing}, pages 5747--5757.

\bibitem[{Hartvigsen et~al.(2022)Hartvigsen, Gabriel, Palangi, Sap, Ray, and
  Kamar}]{hartvigsen2022toxigen}
Thomas Hartvigsen, Saadia Gabriel, Hamid Palangi, Maarten Sap, Dipankar Ray,
  and Ece Kamar. 2022.
\newblock Toxigen: A large-scale machine-generated dataset for adversarial and
  implicit hate speech detection.
\newblock In \emph{Proceedings of the 60th Annual Meeting of the Association
  for Computational Linguistics (Volume 1: Long Papers)}, pages 3309--3326.

\bibitem[{He(2023)}]{gptfast}
Horace He. 2023.
\newblock {GPT-Fast}.
\newblock [Online]. Available:
  \url{https://github.com/pytorch-labs/gpt-fast/tree/main?tab=readme-ov-file}.
\newblock Accessed: Feb. 2, 2024.

\bibitem[{He et~al.(2024)He, Zhong, Cai, Lee, and He}]{he2023rest}
Zhenyu He, Zexuan Zhong, Tianle Cai, Jason Lee, and Di~He. 2024.
\newblock Rest: Retrieval-based speculative decoding.
\newblock In \emph{Proceedings of the 2024 Conference of the North American
  Chapter of the Association for Computational Linguistics: Human Language
  Technologies (Volume 1: Long Papers)}, pages 1582--1595.

\bibitem[{Hendrycks et~al.(2020)Hendrycks, Burns, Basart, Zou, Mazeika, Song,
  and Steinhardt}]{hendrycks2020measuring}
Dan Hendrycks, Collin Burns, Steven Basart, Andy Zou, Mantas Mazeika, Dawn
  Song, and Jacob Steinhardt. 2020.
\newblock Measuring massive multitask language understanding.
\newblock In \emph{International Conference on Learning Representations}.

\bibitem[{Jang et~al.(2016)Jang, Gu, and Poole}]{jang2016categorical}
Eric Jang, Shixiang Gu, and Ben Poole. 2016.
\newblock Categorical reparameterization with gumbel-softmax.
\newblock In \emph{International Conference on Learning Representations}.

\bibitem[{Jiang et~al.(2023)Jiang, Sablayrolles, Mensch, Bamford, Chaplot,
  Casas, Bressand, Lengyel, Lample, Saulnier et~al.}]{jiang2023mistral}
Albert~Q Jiang, Alexandre Sablayrolles, Arthur Mensch, Chris Bamford,
  Devendra~Singh Chaplot, Diego de~las Casas, Florian Bressand, Gianna Lengyel,
  Guillaume Lample, Lucile Saulnier, et~al. 2023.
\newblock Mistral 7b.
\newblock \emph{arXiv preprint arXiv:2310.06825}.

\bibitem[{Kaddour et~al.(2023)Kaddour, Harris, Mozes, Bradley, Raileanu, and
  McHardy}]{kaddour2023challenges}
Jean Kaddour, Joshua Harris, Maximilian Mozes, Herbie Bradley, Roberta
  Raileanu, and Robert McHardy. 2023.
\newblock Challenges and applications of large language models.
\newblock \emph{arXiv preprint arXiv:2307.10169}.

\bibitem[{Kendall(1938)}]{kendall1938new}
Maurice~G Kendall. 1938.
\newblock A new measure of rank correlation.
\newblock \emph{Biometrika}, 30(1/2):81--93.

\bibitem[{Lermen and Rogers-Smith(2024)}]{lermen2023lora}
Simon Lermen and Charlie Rogers-Smith. 2024.
\newblock Lora fine-tuning efficiently undoes safety training in llama 2-chat
  70b.
\newblock In \emph{ICLR 2024 Workshop on Secure and Trustworthy Large Language
  Models}.

\bibitem[{Lester et~al.(2021)Lester, Al-Rfou, and Constant}]{lester2021power}
Brian Lester, Rami Al-Rfou, and Noah Constant. 2021.
\newblock The power of scale for parameter-efficient prompt tuning.
\newblock In \emph{Proceedings of the 2021 Conference on Empirical Methods in
  Natural Language Processing}, pages 3045--3059.

\bibitem[{Leviathan et~al.(2023)Leviathan, Kalman, and
  Matias}]{leviathan2023fast}
Yaniv Leviathan, Matan Kalman, and Yossi Matias. 2023.
\newblock Fast inference from transformers via speculative decoding.
\newblock In \emph{International Conference on Machine Learning}, pages
  19274--19286. PMLR.

\bibitem[{Lewis et~al.(2019)Lewis, Liu, Goyal, Ghazvininejad, Mohamed, Levy,
  Stoyanov, and Zettlemoyer}]{lewis2019bart}
Mike Lewis, Yinhan Liu, Naman Goyal, Marjan Ghazvininejad, Abdelrahman Mohamed,
  Omer Levy, Ves Stoyanov, and Luke Zettlemoyer. 2019.
\newblock Bart: Denoising sequence-to-sequence pre-training for natural
  language generation, translation, and comprehension.
\newblock \emph{arXiv preprint arXiv:1910.13461}.

\bibitem[{Li et~al.(2023)Li, Bubeck, Eldan, Del~Giorno, Gunasekar, and
  Lee}]{textbooks2}
Yuanzhi Li, S{\'e}bastien Bubeck, Ronen Eldan, Allie Del~Giorno, Suriya
  Gunasekar, and Yin~Tat Lee. 2023.
\newblock Textbooks are all you need ii: \textbf{phi-1.5} technical report.
\newblock \emph{arXiv preprint arXiv:2309.05463}.

\bibitem[{Liu et~al.(2021)Liu, Dai, So, and Le}]{liu2021pay}
Hanxiao Liu, Zihang Dai, David So, and Quoc~V Le. 2021.
\newblock Pay attention to mlps.
\newblock \emph{Advances in Neural Information Processing Systems},
  34:9204--9215.

\bibitem[{Liu et~al.(2024)Liu, Xu, Chen, and Xiao}]{liu2023autodan}
Xiaogeng Liu, Nan Xu, Muhao Chen, and Chaowei Xiao. 2024.
\newblock Autodan: Generating stealthy jailbreak prompts on aligned large
  language models.
\newblock In \emph{The Twelfth International Conference on Learning
  Representations}.

\bibitem[{Liu et~al.(2023)Liu, Yao, Ton, Zhang, Guo, Cheng, Klochkov, Taufiq,
  and Li}]{liu2023trustworthy}
Yang Liu, Yuanshun Yao, Jean-Francois Ton, Xiaoying Zhang, Ruocheng Guo, Hao
  Cheng, Yegor Klochkov, Muhammad~Faaiz Taufiq, and Hang Li. 2023.
\newblock Trustworthy llms: a survey and guideline for evaluating large
  language models' alignment.
\newblock In \emph{Socially Responsible Language Modelling Research}.

\bibitem[{Liu et~al.(2019)Liu, Ott, Goyal, Du, Joshi, Chen, Levy, Lewis,
  Zettlemoyer, and Stoyanov}]{liu2019roberta}
Yinhan Liu, Myle Ott, Naman Goyal, Jingfei Du, Mandar Joshi, Danqi Chen, Omer
  Levy, Mike Lewis, Luke Zettlemoyer, and Veselin Stoyanov. 2019.
\newblock Roberta: A robustly optimized bert pretraining approach.
\newblock \emph{arXiv preprint arXiv:1907.11692}.

\bibitem[{Long et~al.(2024)Long, Zhao, Brown, Xie, Zhao, Chen, Kawaguchi,
  Shieh, and He}]{do2023prompt}
Do~Long, Yiran Zhao, Hannah Brown, Yuxi Xie, James Zhao, Nancy Chen, Kenji
  Kawaguchi, Michael Shieh, and Junxian He. 2024.
\newblock \href {https://doi.org/10.18653/v1/2024.acl-long.395} {Prompt
  optimization via adversarial in-context learning}.
\newblock In \emph{Proceedings of the 62nd Annual Meeting of the Association
  for Computational Linguistics (Volume 1: Long Papers)}, pages 7308--7327,
  Bangkok, Thailand. Association for Computational Linguistics.

\bibitem[{Lu et~al.(2023)Lu, Qiu, Chang, Wu, Zhu, Rajpurohit, Clark, and
  Kalyan}]{ludynamic}
Pan Lu, Liang Qiu, Kai-Wei Chang, Ying~Nian Wu, Song-Chun Zhu, Tanmay
  Rajpurohit, Peter Clark, and Ashwin Kalyan. 2023.
\newblock Dynamic prompt learning via policy gradient for semi-structured
  mathematical reasoning.
\newblock In \emph{The Eleventh International Conference on Learning
  Representations}.

\bibitem[{Maddison et~al.(2022)Maddison, Mnih, and Teh}]{maddison2016concrete}
Chris~J Maddison, Andriy Mnih, and Yee~Whye Teh. 2022.
\newblock The concrete distribution: A continuous relaxation of discrete random
  variables.
\newblock In \emph{International Conference on Learning Representations}.

\bibitem[{Marelli et~al.(2014)Marelli, Menini, Baroni, Bentivogli, Bernardi,
  and Zamparelli}]{marelli-etal-2014-sick}
Marco Marelli, Stefano Menini, Marco Baroni, Luisa Bentivogli, Raffaella
  Bernardi, and Roberto Zamparelli. 2014.
\newblock \href
  {http://www.lrec-conf.org/proceedings/lrec2014/pdf/363_Paper.pdf} {A {SICK}
  cure for the evaluation of compositional distributional semantic models}.
\newblock In \emph{Proceedings of the Ninth International Conference on
  Language Resources and Evaluation ({LREC}'14)}, pages 216--223, Reykjavik,
  Iceland. European Language Resources Association (ELRA).

\bibitem[{Mingkai and Jianyu(2022)}]{mingkai2022rlprompt}
Deng Mingkai and Wang Jianyu. 2022.
\newblock Rlprompt: Optimizing discrete text prompts with reinforcement
  learning.
\newblock In \emph{Proceedings of the 2022 Conference on Empirical Methods in
  Natural Language Processing}.

\bibitem[{M{\"o}kander et~al.(2023)M{\"o}kander, Schuett, Kirk, and
  Floridi}]{mokander2023auditing}
Jakob M{\"o}kander, Jonas Schuett, Hannah~Rose Kirk, and Luciano Floridi. 2023.
\newblock Auditing large language models: a three-layered approach.
\newblock \emph{AI and Ethics}, pages 1--31.

\bibitem[{Ouyang et~al.(2022)Ouyang, Wu, Jiang, Almeida, Wainwright, Mishkin,
  Zhang, Agarwal, Slama, Ray et~al.}]{ouyang2022training}
Long Ouyang, Jeffrey Wu, Xu~Jiang, Diogo Almeida, Carroll Wainwright, Pamela
  Mishkin, Chong Zhang, Sandhini Agarwal, Katarina Slama, Alex Ray, et~al.
  2022.
\newblock Training language models to follow instructions with human feedback.
\newblock \emph{Advances in Neural Information Processing Systems},
  35:27730--27744.

\bibitem[{Pearson(1900)}]{pearson1900x}
Karl Pearson. 1900.
\newblock X. on the criterion that a given system of deviations from the
  probable in the case of a correlated system of variables is such that it can
  be reasonably supposed to have arisen from random sampling.
\newblock \emph{The London, Edinburgh, and Dublin Philosophical Magazine and
  Journal of Science}, 50(302):157--175.

\bibitem[{Pincus(1970)}]{pincus1970monte}
Martin Pincus. 1970.
\newblock A monte carlo method for the approximate solution of certain types of
  constrained optimization problems.
\newblock \emph{Operations research}, 18(6):1225--1228.

\bibitem[{Pryzant et~al.(2023)Pryzant, Iter, Li, Lee, Zhu, and
  Zeng}]{pryzant2023automatic}
Reid Pryzant, Dan Iter, Jerry Li, Yin Lee, Chenguang Zhu, and Michael Zeng.
  2023.
\newblock Automatic prompt optimization with “gradient descent” and beam
  search.
\newblock In \emph{Proceedings of the 2023 Conference on Empirical Methods in
  Natural Language Processing}, pages 7957--7968.

\bibitem[{Qi et~al.(2024)Qi, Zeng, Xie, Chen, Jia, Mittal, and
  Henderson}]{qi2023fine}
Xiangyu Qi, Yi~Zeng, Tinghao Xie, Pin-Yu Chen, Ruoxi Jia, Prateek Mittal, and
  Peter Henderson. 2024.
\newblock Fine-tuning aligned language models compromises safety, even when
  users do not intend to!
\newblock In \emph{The Twelfth International Conference on Learning
  Representations}.

\bibitem[{Radford et~al.(2019)Radford, Wu, Child, Luan, Amodei, Sutskever
  et~al.}]{radford2019language}
Alec Radford, Jeffrey Wu, Rewon Child, David Luan, Dario Amodei, Ilya
  Sutskever, et~al. 2019.
\newblock Language models are unsupervised multitask learners.

\bibitem[{Rando and Tram{\`e}r(2023)}]{rando2023universal}
Javier Rando and Florian Tram{\`e}r. 2023.
\newblock Universal jailbreak backdoors from poisoned human feedback.
\newblock \emph{arXiv preprint arXiv:2311.14455}.

\bibitem[{Shayegani et~al.(2024)Shayegani, Dong, and
  Abu-Ghazaleh}]{shayegani2023jailbreak}
Erfan Shayegani, Yue Dong, and Nael Abu-Ghazaleh. 2024.
\newblock Jailbreak in pieces: Compositional adversarial attacks on multi-modal
  language models.
\newblock In \emph{The Twelfth International Conference on Learning
  Representations}.

\bibitem[{Shin et~al.(2020)Shin, Razeghi, Logan~IV, Wallace, and
  Singh}]{shin2020autoprompt}
Taylor Shin, Yasaman Razeghi, Robert~L Logan~IV, Eric Wallace, and Sameer
  Singh. 2020.
\newblock Autoprompt: Eliciting knowledge from language models with
  automatically generated prompts.
\newblock In \emph{Proceedings of the 2020 Conference on Empirical Methods in
  Natural Language Processing (EMNLP)}, pages 4222--4235.

\bibitem[{So et~al.(2019)So, Le, and Liang}]{so2019evolved}
David So, Quoc Le, and Chen Liang. 2019.
\newblock The evolved transformer.
\newblock In \emph{International conference on machine learning}, pages
  5877--5886. PMLR.

\bibitem[{Socher et~al.(2013)Socher, Perelygin, Wu, Chuang, Manning, Ng, and
  Potts}]{socher2013recursive}
Richard Socher, Alex Perelygin, Jean Wu, Jason Chuang, Christopher~D Manning,
  Andrew~Y Ng, and Christopher Potts. 2013.
\newblock Recursive deep models for semantic compositionality over a sentiment
  treebank.
\newblock In \emph{Proceedings of the 2013 conference on empirical methods in
  natural language processing}, pages 1631--1642.

\bibitem[{Stiennon et~al.(2020)Stiennon, Ouyang, Wu, Ziegler, Lowe, Voss,
  Radford, Amodei, and Christiano}]{stiennon2020learning}
Nisan Stiennon, Long Ouyang, Jeffrey Wu, Daniel Ziegler, Ryan Lowe, Chelsea
  Voss, Alec Radford, Dario Amodei, and Paul~F Christiano. 2020.
\newblock Learning to summarize with human feedback.
\newblock \emph{Advances in Neural Information Processing Systems},
  33:3008--3021.

\bibitem[{Suzgun et~al.(2023)Suzgun, Scales, Sch{\"a}rli, Gehrmann, Tay, Chung,
  Chowdhery, Le, Chi, Zhou et~al.}]{suzgun2023challenging}
Mirac Suzgun, Nathan Scales, Nathanael Sch{\"a}rli, Sebastian Gehrmann, Yi~Tay,
  Hyung~Won Chung, Aakanksha Chowdhery, Quoc Le, Ed~Chi, Denny Zhou, et~al.
  2023.
\newblock Challenging big-bench tasks and whether chain-of-thought can solve
  them.
\newblock In \emph{Findings of the Association for Computational Linguistics:
  ACL 2023}, pages 13003--13051.

\bibitem[{Touvron et~al.(2023)Touvron, Martin, Stone, Albert, Almahairi,
  Babaei, Bashlykov, Batra, Bhargava, Bhosale et~al.}]{touvron2023llama}
Hugo Touvron, Louis Martin, Kevin Stone, Peter Albert, Amjad Almahairi, Yasmine
  Babaei, Nikolay Bashlykov, Soumya Batra, Prajjwal Bhargava, Shruti Bhosale,
  et~al. 2023.
\newblock Llama 2: Open foundation and fine-tuned chat models.
\newblock \emph{arXiv preprint arXiv:2307.09288}.

\bibitem[{Toyer et~al.(2023)Toyer, Watkins, Mendes, Svegliato, Bailey, Wang,
  Ong, Elmaaroufi, Abbeel, Darrell et~al.}]{toyer2023tensor}
Sam Toyer, Olivia Watkins, Ethan Mendes, Justin Svegliato, Luke Bailey, Tiffany
  Wang, Isaac Ong, Karim Elmaaroufi, Pieter Abbeel, Trevor Darrell, et~al.
  2023.
\newblock Tensor trust: Interpretable prompt injection attacks from an online
  game.
\newblock In \emph{NeurIPS 2023 Workshop on Instruction Tuning and Instruction
  Following}.

\bibitem[{Wen et~al.(2024)Wen, Jain, Kirchenbauer, Goldblum, Geiping, and
  Goldstein}]{wen2023hard}
Yuxin Wen, Neel Jain, John Kirchenbauer, Micah Goldblum, Jonas Geiping, and Tom
  Goldstein. 2024.
\newblock Hard prompts made easy: Gradient-based discrete optimization for
  prompt tuning and discovery.
\newblock \emph{Advances in Neural Information Processing Systems}, 36.

\bibitem[{Xia et~al.(2023)Xia, Gao, Zeng, and Chen}]{xia2023sheared}
Mengzhou Xia, Tianyu Gao, Zhiyuan Zeng, and Danqi Chen. 2023.
\newblock Sheared llama: Accelerating language model pre-training via
  structured pruning.
\newblock In \emph{Workshop on Advancing Neural Network Training: Computational
  Efficiency, Scalability, and Resource Optimization (WANT@ NeurIPS 2023)}.

\bibitem[{Xu et~al.(2023)Xu, Chern, Chern, Zhang, Wang, Liu, Li, Fu, and
  Liu}]{xu2023align}
Chunpu Xu, Steffi Chern, Ethan Chern, Ge~Zhang, Zekun Wang, Ruibo Liu, Jing Li,
  Jie Fu, and Pengfei Liu. 2023.
\newblock Align on the fly: Adapting chatbot behavior to established norms.
\newblock \emph{arXiv preprint arXiv:2312.15907}.

\bibitem[{Xu et~al.(2022)Xu, Chen, Du, Shao, Yanggang, Li, and
  Yang}]{xu2022gps}
Hanwei Xu, Yujun Chen, Yulun Du, Nan Shao, Wang Yanggang, Haiyu Li, and Zhilin
  Yang. 2022.
\newblock Gps: Genetic prompt search for efficient few-shot learning.
\newblock In \emph{Proceedings of the 2022 Conference on Empirical Methods in
  Natural Language Processing}, pages 8162--8171.

\bibitem[{Yuan et~al.(2024{\natexlab{a}})Yuan, Pang, Cho, Li, Sukhbaatar, Xu,
  and Weston}]{yuan2024self}
Weizhe Yuan, Richard~Yuanzhe Pang, Kyunghyun Cho, Xian Li, Sainbayar
  Sukhbaatar, Jing Xu, and Jason~E Weston. 2024{\natexlab{a}}.
\newblock Self-rewarding language models.
\newblock In \emph{Forty-first International Conference on Machine Learning}.

\bibitem[{Yuan et~al.(2024{\natexlab{b}})Yuan, Jiao, Wang, Huang, He, Shi, and
  Tu}]{yuan2023gpt}
Youliang Yuan, Wenxiang Jiao, Wenxuan Wang, Jen-tse Huang, Pinjia He, Shuming
  Shi, and Zhaopeng Tu. 2024{\natexlab{b}}.
\newblock Gpt-4 is too smart to be safe: Stealthy chat with llms via cipher.
\newblock In \emph{The Twelfth International Conference on Learning
  Representations}.

\bibitem[{Zar(2005)}]{zar2005spearman}
Jerrold~H Zar. 2005.
\newblock Spearman rank correlation.
\newblock \emph{Encyclopedia of biostatistics}, 7.

\bibitem[{Zhang et~al.(2024)Zhang, Zeng, Wang, and Lu}]{zhang2024tinyllama}
Peiyuan Zhang, Guangtao Zeng, Tianduo Wang, and Wei Lu. 2024.
\newblock Tinyllama: An open-source small language model.
\newblock \emph{arXiv preprint arXiv:2401.02385}.

\bibitem[{Zheng et~al.(2023)Zheng, Chiang, Sheng, Zhuang, Wu, Zhuang, Lin, Li,
  Li, Xing et~al.}]{zheng2023judging}
Lianmin Zheng, Wei-Lin Chiang, Ying Sheng, Siyuan Zhuang, Zhanghao Wu, Yonghao
  Zhuang, Zi~Lin, Zhuohan Li, Dacheng Li, Eric Xing, et~al. 2023.
\newblock Judging llm-as-a-judge with mt-bench and chatbot arena.
\newblock \emph{Advances in Neural Information Processing Systems},
  36:46595--46623.

\bibitem[{Zhou et~al.(2022)Zhou, Muresanu, Han, Paster, Pitis, Chan, and
  Ba}]{zhou2022large}
Yongchao Zhou, Andrei~Ioan Muresanu, Ziwen Han, Keiran Paster, Silviu Pitis,
  Harris Chan, and Jimmy Ba. 2022.
\newblock Large language models are human-level prompt engineers.
\newblock In \emph{The Eleventh International Conference on Learning
  Representations}.

\bibitem[{Zou et~al.(2023)Zou, Wang, Kolter, and Fredrikson}]{zou2023universal}
Andy Zou, Zifan Wang, J~Zico Kolter, and Matt Fredrikson. 2023.
\newblock Universal and transferable adversarial attacks on aligned language
  models.
\newblock \emph{arXiv preprint arXiv:2307.15043}.

\end{thebibliography}
\bibliographystyle{acl}

\clearpage

\appendix

\section{Implementation}\label{sec:appen_imple}

The following code shows the core implementation of probe sampling using PyTorch. As seen in the code, the algorithm is relatively easy to use. 

\vspace{-0.3cm}

\begin{minted}[linenos=false, frame=lines, framesep=2mm, fontsize=\scriptsize, fontfamily=courier, breaklines=false]{python}
def draft_model_all(args):
    draft_model.loss(control_cands)
    
    queue.put('draft':loss_small)

def target_model_probe(args):
    probe_index = random.sample(range(512), 512/16)
    probe_control_cands = control_cands[probe_index]
    target_model.loss(probe_control_cands)
    
    queue.put('target':[loss_large_probe, probe_index])

# Parallelly Calculate Loss on Batch and Probe Set
args=(control_cands, batch_size, queue)
threading.Thread(target=draft_model_all, args=args)
threading.Thread(target=target_model_probe, args=args)

# Calculate Agreement Score
cor = spearmanr(loss_small[probe_index], large_loss_probe)

# Target Model Test on Filtered Set
filtered_size = int((1 - cor) * 512/8)
indices = topk(loss_small, k=filtered_size, largest=False)
filtered_control_cands = control_cands[indices]
target_model.loss(filtered_control_cands)

# Return Lowest Loss Candidate
return [large_loss_probe, filtered_control_cands].lowest()
\end{minted}

\section{Converge Process}\label{sec:appen_converge}

\begin{figure}[ht]
  \centering
    \includegraphics[width=0.93\textwidth]{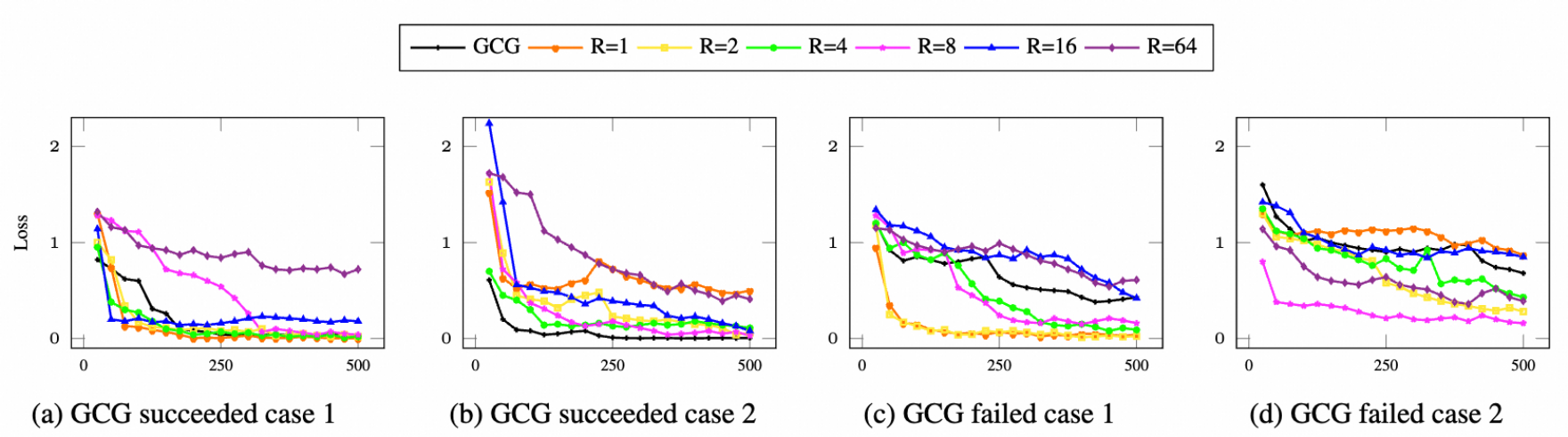}
\caption{Converge progress with different sizes of filtered set.}
\label{fig:fig_r}
\vspace{-0.2cm}
\end{figure}

In Figure \ref{fig:fig_r}, we also show a few convergence processes with different values of $R$, where the pink line corresponds to $R=8$. The pink line always achieves successful optimization while the other lines can lead to suboptimal results due to excessive randomness or insufficient randomness. In particular, the blue and yellow lines can suffer from excessive randomness and the other lines might have insufficient randomness. 

\section{Software optimization}
\begin{wraptable}{r}{0.5\textwidth}
\vspace{-0.4cm}
  \centering
  \caption{Results with torch.compile() enabled. torch.comple() does not lead to further speedup. }
  \vspace{-0.15cm}
  \footnotesize
  \scalebox{0.9}{
    \begin{tabular}{l|l|l|l}
      \toprule
      \textbf{Method} & GCG  & Probe sampling & PS (Compile)   \\
      \midrule
      ASR & $66.0$ & $85.0$ & $85.0$ \\\midrule
      Time (s) & $9.16$ & $2.60\;(3.5\times)$ & $2.54\;(3.6\times)$   \\
      \bottomrule
    \end{tabular}
  }
  \label{table:compile}
\vspace{-0.4cm}
\end{wraptable}

In other speedup works~\citep{gptfast}, using \textit{torch.compile()} can lead to significant acceleration. It compiles LLMs into an kernel and alleviate the overhead of repeatedly launching the kernel. Table \ref{table:compile} shows that the time cost is similar with or without this optimization enabled. This is likely due to the fact that we use large batch sizes and long input sequences, whose computation cost dominates the overhead caused by the eager execution and launching the kernel repeatedly.

\end{document}